\documentclass[10pt,twocolumn,letterpaper]{article}

\usepackage{cvpr}              

\usepackage[dvipsnames]{xcolor}

\usepackage{multirow}
\usepackage{svg}
\usepackage{soul}
\usepackage[symbol]{footmisc}
\usepackage{gensymb}

\definecolor{cvprblue}{rgb}{0.21,0.49,0.74}
\usepackage[pagebackref,breaklinks,colorlinks,citecolor=cvprblue]{hyperref}
\usepackage{svg}
\usepackage[accsupp]{axessibility}

\def\paperID{3} 
\def\confName{CVPR}
\def\confYear{2024}

\title{Faster Than Lies: Real-time Deepfake Detection using Binary Neural Networks}

\author{Romeo Lanzino, Federico Fontana, Anxhelo Diko, Marco Raoul Marini, Luigi Cinque\\
Department of Computer Science\\Sapienza University of Rome\\Via Salaria 113, 00198\\
Rome, Italy\\
{\tt\small \{lanzino, fontana.f, diko, marini, cinque\}@di.uniroma1.it}
}

\begin{document}
\maketitle
\begin{abstract}

Deepfake detection aims to contrast the spread of deep-generated media that undermines trust in online content. While existing methods focus on large and complex models, the need for real-time detection demands greater efficiency. With this in mind, unlike previous work, we introduce a novel deepfake detection approach on images using Binary Neural Networks (BNNs) for fast inference with minimal accuracy loss. Moreover, our method incorporates Fast Fourier Transform (FFT) and Local Binary Pattern (LBP) as additional channel features to uncover manipulation traces in frequency and texture domains. Evaluations on COCOFake, DFFD, and CIFAKE datasets demonstrate our method's state-of-the-art performance in most scenarios with a significant efficiency gain of up to a $20\times$ reduction in FLOPs during inference. Finally, by exploring BNNs in deepfake detection to balance accuracy and efficiency, this work paves the way for future research on efficient deepfake detection.
\end{abstract}  
\section{Introduction}
\label{sec:intro}

The rise of deepfakes and media manipulated with sophisticated artificial intelligence threatens to erode the foundation of trust in the digital age.  From fabricated revenge pornography \cite{chesney2019deep} to doctored political speeches \cite{kietzmann2020deepfakes}, these synthetic creations have the power to deceive, defame, and destabilize. As deepfake creation tools become more accessible and the quality of these fakes rapidly improves \cite{li2020celeb,coco_fake}, the ability to distinguish authentic content from malicious manipulation has become a critical battleground for preserving truth and preventing the diffusion of fake information. This urge to identify deepfakes has brought much attention to the rapidly evolving field of deepfakes detection \cite{ciftci2020fakecatcher,li2018ictu,guarnera2020deepfake,afchar2018mesonet}. As the name suggests, this field concentrates on creating intelligent algorithms that are able to depict common patterns that distinguish real from fake content \cite{coco_fake}.

\begin{figure}[t]
\includegraphics[width=\linewidth]{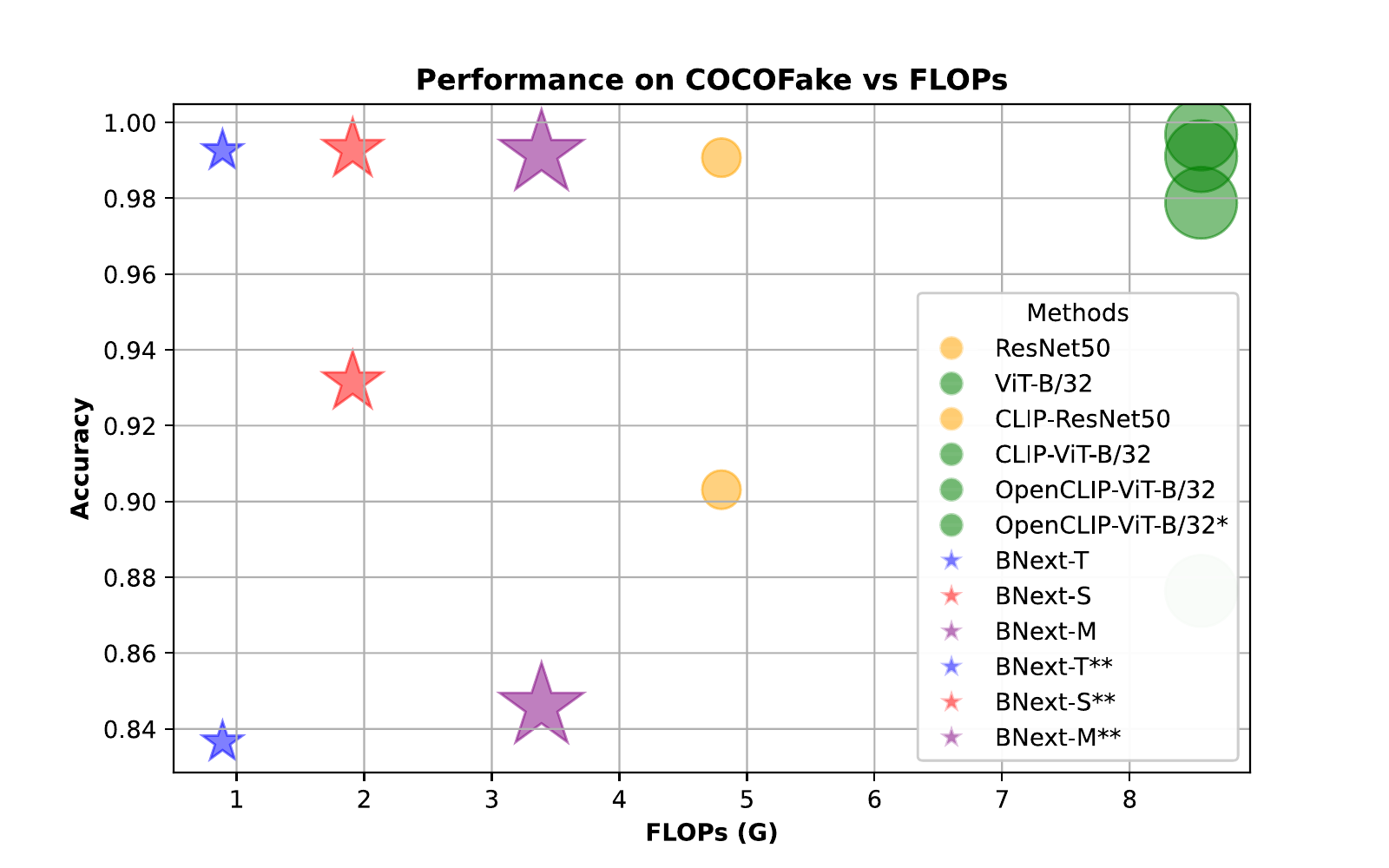}
 \caption{Depiction of the trade-off between performance (accuracy) and the computational complexity measured in FLOPs (G) on the COCOFake \cite{coco_fake} dataset. Points with the same color indicate models that share the same architecture. The size of a point represents the number of parameters of its model. $^\ast$This model shares the architecture with the one with the same name but is trained on a different dataset. $^{\ast\ast}$In these models, which are the ones on the bottom left part, the backbone is kept frozen.}
 \label{fig:teaser}
\end{figure}

Deepfake detection is an evolving task driven by the relentless advancement of generative models, especially in image generation. In such regard, recent years have seen remarkable progress in generating realistic images, thanks to Generative Adversarial Networks (GANs) \cite{goodfellow2016deep}. This spurred the creation of deepfake detection algorithms, often based on deep neural networks \cite{ciftci2020fakecatcher,afchar2018mesonet,faceforensics}, that accurately identified characteristic GAN-based artifacts. However, the emergence of diffusion models \cite{hu2017diffusion,stable_diffusion,coco_fake}, capable of even higher-quality fakes, challenges these methods \cite{nichol2021glide,ramesh2022hierarchical,saharia2022photorealistic}. In tackling the challenges of identifying fake images generated by diffusion models, previous works concentrate on training large neural networks with millions of parameters to identify the inherent patterns of generated images \cite{coco_fake}. While these detection methods show promise, their reliance on large, complex models raises concerns.  Deepfakes primarily spread on social media platforms and web applications \cite{kalpokas2022deepfakes}, where devices like mobile phones and personal computers have limited computational resources. This raises a critical question: \textit{Can we develop efficient deep-learning-based deepfake detection methods without sacrificing accuracy?}

In the pursuit of answering our question, we shift the attention to Binary Neural Networks (BNNs). BNNs offer exceptional memory and computational savings by quantizing both weights and activations to 1-bit values \cite{han2020training,rastegari2016xnor,xu2021learning,yang2020searching}. This makes them ideal for real-time deepfake detection on resource-constrained devices like phones and personal computers. Specifically, we employ the BNext \cite{guo2022join} convolutional neural network for its proven feature extraction capabilities on RGB images. Additionally, since generative methods often leave subtle artifacts, particularly around edges and in frequency domains \cite{li2020celeb,yang2019exposing,guarnera2020deepfake}, we augment our RGB input with features derived from two specialized filters: the Fast Fourier Transform (FFT) magnitude, and the Local Binary Patterns (LBP). Such features further enhance the model's ability to identify generated artifacts by emphasizing micro-patterns and textures. 

To assess the performance of our method's ability to identify deep-generated content and its efficiency in terms of computational resources, we conduct extensive experiments in three deepfakes detection datasets, including COCOFake \cite{coco_fake}, DFFD \cite{dffd}, and CIFAKE \cite{cifake}. Our method competes or improves the results of existing SOTA in almost all scenarios while reducing up to $20\times$ in computational consumption measured in FLOPs. A glimpse of the performance and computational complexity trade-off of our method compared to the current SOTA can be seen in Fig. \ref{fig:teaser}.

In summary, the contributions of this work are four-fold:
\begin{enumerate}
    \item to the best of our knowledge, we propose the first-ever implementation of a BNN for deepfake detection, which enables real-time detection on low-resource devices;
    \item extensive experimentation of three benchmark datasets (COCOFake, DFFD, and CIFAKE) showcasing the exceptional ability of BNNs in identifying generated images;
    \item an ablation study that highlights the impact of each design choice made in building the proposed method;
    \item quantitative results that underline the quality of the proposal and promote further investigation.
\end{enumerate}

The remaining parts of this work are structured as follows: Section \ref{sec:related_work} delves into the existing literature, situating our work within the broader context of deepfake detection and BNN advancements; Section \ref{sec:method} details the methodology employed, including our use of BNNs and the rationale behind augmenting input images with FFT magnitude and LBP channels for enhanced detection capabilities; Section \ref{sec:experiments} presents the experimental setup, the datasets utilized, the metrics for evaluation, and the results obtained, alongside an ablation study to discern the impact of each augmentation of the input images; finally, Section \ref{sec:conclusion} summarizes the findings and the limitations of our work on the deepfake detection task, outlining avenues for future research.

The code is available at \url{https://github.com/fedeloper/binary_deepfake_detection}.

\section{Related work}
\label{sec:related_work}

\subsection{Deepfakes detection}
The fight against deepfakes has spurred intense research efforts, yielding a range of detection strategies. Early methods focused on low-level artifacts stemming from generative processes, pinpointing anomalies like unnatural blinking patterns \cite{li2018ictu} or inconsistencies in physiological signals patterns \cite{ciftci2020fakecatcher}. However, these techniques are vulnerable to increasingly sophisticated deepfake generation methods. A more robust approach lies in analyzing mesoscopic features, such as facial warping artifacts \cite{li2020celeb}, inconsistencies in head pose \cite{yang2019exposing}, and textural anomalies \cite{guarnera2020deepfake}. Deep neural networks, specifically CNNs, have emerged as powerful detection tools \cite{afchar2018mesonet}. Subsequent works have explored tailored architectures like XceptionNet \cite{faceforensics} and attention mechanisms \cite{nguyen2019multi} to enhance artifact detection. To address the data scarcity issue and improve adaptability to new deepfake techniques, self-supervised \cite{hsu2020deep} and semi-supervised methods \cite{cozzolino2018forensictransfer}  are gaining traction. 
However, current techniques primarily detect GAN-generated samples, limiting their effectiveness across the full range of deepfake creation methods \cite{coco_fake}. In recent years, diffusion models introduced new milestones in deepfake generation with high-quality images resembling natural ones \cite{stable_diffusion}. The detection of such models' images presents more challenges than those created with traditional GANs, often avoiding the telltale grid-like artifacts found in GAN outputs and requiring a shift in detection strategies \cite{nichol2021glide,ramesh2022hierarchical,saharia2022photorealistic}.  Promising research focuses on analyzing the intrinsic local dimensionality of diffusion-generated images \cite{lorenz2023detecting}, which differs from natural images. Another approach investigates how diffusion models tend to overfit training data, leaving detectable traces in the form of reconstruction errors \cite{wang2020cnn}. While tailored methods are emerging, it is important to note that most works rely on the use of advanced neural networks like heavy CNNs or Transformers \cite{coco_fake} to detect deepfakes. Despite notable progress, deepfake detection faces ongoing challenges.  
Models often struggle to generalize to unseen deepfake generation techniques, and their performance degrades when encountering real-world distortions (e.g., video compression \cite{guarnera2020deepfake}). 

\subsection{Binary Neural Networks}
The BNN architecture, pioneered by \cite{courbariaux2016binarized}, involves binarizing weights and activations through the \textit{sign} function, substituting the bulk of arithmetic operations in deep neural networks with bit-wise operations.
To address quantization error, the XOR-Net \cite{rastegari2016xnor} introduced a channel-wise scaling factor for reconstructing binarized weights, a technique pivotal in subsequent BNN models. As proposed in \cite{lin2017towards}, ABC-Net endeavors to approximate full-precision weights with a linear combination of binary weight bases and employs multiple binary activations to diminish information loss.
Inspired by the full-precision networks ResNet \cite{he2016deep} and DenseNet \cite{huang2017densely} architectures, Bi-Real Net \cite{liu2018bi} integrates shortcuts to minimize the performance gap between 1-bit and real-valued CNN models. 
Concurrently, BinaryDenseNet \cite{bethge2019back} enhances BNN accuracy by increasing the number of shortcuts. 
Further, IR-Net \cite{qin2020forward} proposes the Libra-PB method, aimed at reducing information loss during forward propagation through the maximization of quantized parameters' information entropy and minimizing quantization error, bounded within $\{-1, +1\}$. ReActNet \cite{liu2020reactnet} develops a generalized version of traditional \textit{sign} and \textit{PReLU} functions, named \textit{RSign} and \textit{RPReLU}, respectively, facilitating explicit learning of distribution reshaping and shifting with minimal computational overhead.
RBNN \cite{lin2020rotated} examines and mitigates angular bias's effect on quantization error. 
SiMaN \cite{lin2021siman} reveals that removing $L_2$ Regularization during training maximizes the entropy. 
ReCU~\cite{xu2021recu} introduces a rectified clamp unit to revive the so-called "dead weights", thus reducing quantization error.
AdaBin \cite{tu2022adabin} integrates equalization methods for weights and introduces learnable parameters for activations, employing the \textit{MaxOut} nonlinear activation function to add a negligible count of floating-point operations. 
B-Next \cite{guo2022join} proposed a hybrid approach with a basic block with binarized convolutions and INT-4 linear layers, achieving state-of-the-art performances.
    
\begin{figure*}[!ht]
    \centering

    \includegraphics[width=\textwidth]{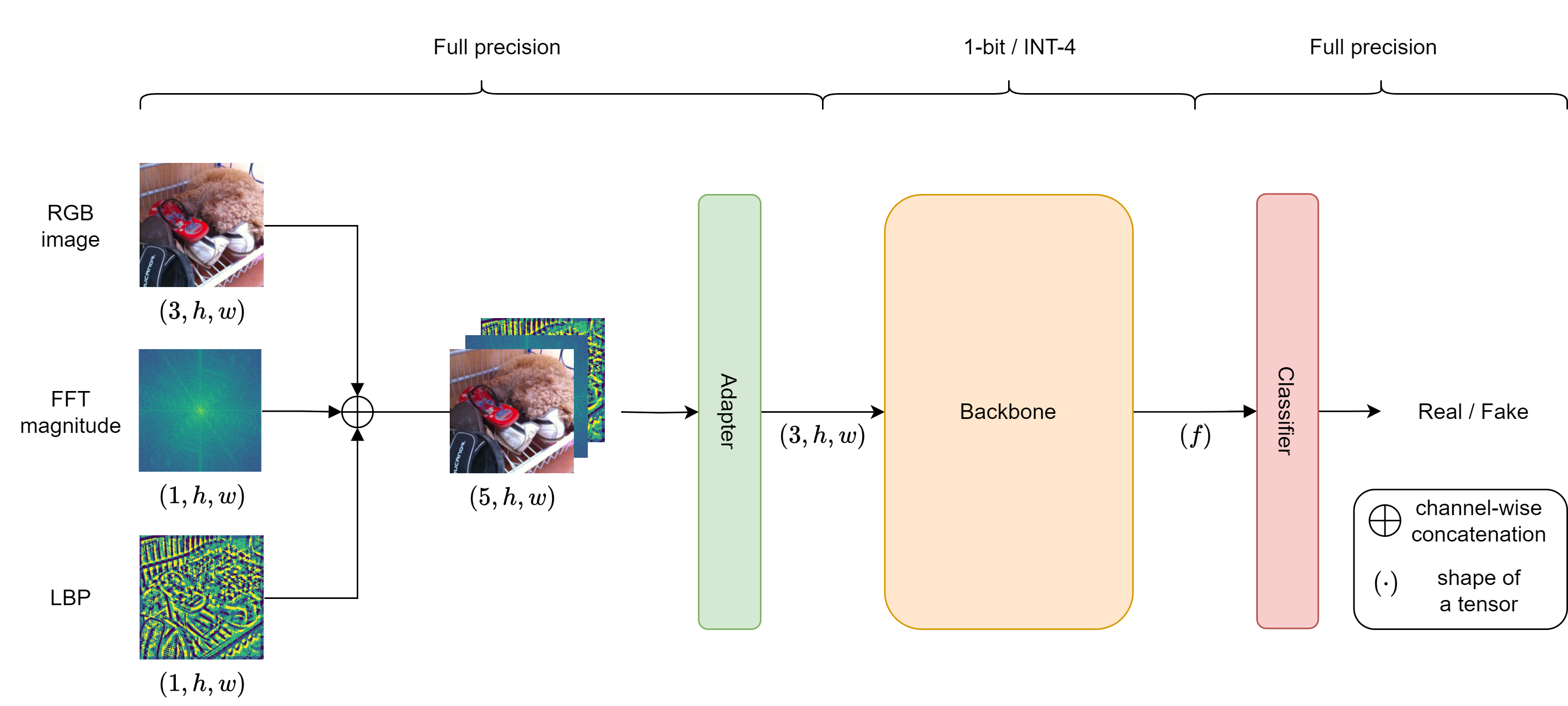}
    \caption{The architecture of the proposed model.}
    \label{fig:architecture}
\end{figure*}

\section{Proposed method}
\label{sec:method}

The proposed method, whose architecture is shown in Figure \ref{fig:architecture}, processes an RGB image to classify it as either real or generated. 
Initially, the method augments the input image by adding two additional channels that correspond to the FFT magnitude and the LBP \cite{lbp}. 
Subsequently, these augmented images undergo adaptation through an Adapter to revert them to a 3-channel format, which is then fed into the backbone for feature extraction. The extracted features are then classified as real or fake. 

\subsection{Augmented features}
\label{sec:augmented_features}
The model takes an RGB image in $\mathbb{R}^{3 \times h \times w}$ as input, where $h$ and $w$ are its height and width, respectively.
This study enriches this image with two channels representing its FFT magnitude and LBP. These augmentations are specifically selected to underscore the subtle yet significant micro-patterns that deepfakes often disrupt, leveraging the intuition that specific texture and edge information can be pivotal in distinguishing between genuine and generated imagery.


We assume that deepfakes could introduce distortions in the frequency domain which are typically not present in genuine images. Thus, we exploited the FFT magnitude channel to highlight these anomalies. This channel is obtained by applying the FFT to the image to extract its magnitude spectrum.

The LBP is a texture descriptor that encapsulates the local spatial structure of an image. It is introduced in the pipeline to capture the unique textures of facial features. Those are areas where deepfakes typically struggle to maintain accuracy. LBP enriches the model's input with robust texture pattern features by comparing each pixel with its neighbors and encoding this comparison into a new image.

\subsection{Adapter}
 
As the backbone is compatible with 3-channel images and the augmented image has more than those channels, we introduced a particular layer just before the backbone, namely the Adapter, to manage the new features.
It is a convolution layer that takes as input an image with $5$ channels (red, green, blue, FFT magnitude, and LBP) and squeezes them into an image of the same height and width but with just $3$ channels, according to the shape of the input accepted by the backbone.

\subsection{Binary backbone}

The concept of a BNN was pioneered by~\cite{courbariaux2016binarized}. Their innovative approach entailed using the sign function to binarize weights and activations, effectively replacing the majority of arithmetic computations in deep neural networks with bit-wise operations, obtaining a theoretical speedup of $58 \times$ in inference speed and $32 \times$ times less memory needed.
Before the explanation of our proposed BNN for this work, we first describe what are the differences between a full-precision and a binary CNN.

We describe a CNN by representing its layer-specific real-valued weights as $\mathcal{W}_r$ and the inputs as $\mathcal{A}_r$. Consequently, the output $\mathcal{Y}$ of a convolution is formulated as follows:
\begin{equation}
\mathcal{Y} = \mathcal{A}_r \otimes \mathcal{W}_r ,
\end{equation}
with $\otimes$ symbolizing standard convolution operations. 
BNNs seeks to convert every weight in $ \mathcal{W}_r$ and every activation $\mathcal{A}_r$ into their binary counterparts $\mathcal{W}_b$ and $\mathcal{A}_b$, whose values are quantized in $\{-1, +1\}$ using the $\textit{sign}$ function. 
For a real-valued input $x_r$, this function is defined as follows:
\begin{equation}
\textit{sign}(x_r)=
\begin{cases}
    +1, & \text{if } x_r \ge 0, \\
    -1, & \text{otherwise}.
\end{cases}
\end{equation}
To address the significant quantization error inherent in deep neural networks binarization, XNOR-Net \cite{courbariaux2016binarized} introduces dual scaling factors for both weights $\mathcal{W}_b$ and activations $\mathcal{A}_b$. Following the methodology outlined by \cite{courbariaux2016binarized}, this document simplifies the representation of these scaling factors to a single parameter $\alpha$. 
Hence, the output of a binary convolution is expressed as follows:
\begin{equation}
\mathcal{Y}  = \mathcal{A}_r \otimes \mathcal{W}_r \approx (\mathcal{A}_b \circledast \mathcal{W}_b) \odot \alpha,
\end{equation}
where $\circledast$ denotes bit-wise operations including XNOR and POPCOUNT, and $\odot$ stands for element-wise multiplication.

The proposed method uses BNext \cite{guo2022join} as a backbone, which is pre-trained on the ImageNet \cite{krizhevsky2012imagenet} dataset. In BNext, each convolution operation is binary, and other operations are quantized to INT-4, resulting in much more efficient networks in terms of operations. 
The backbone uses a binary convolution module with full precision skip-connection and a branch with precision INT-4 to facilitate information propagation and alleviate possible bottlenecks.

The $f$ features extracted by the backbone, represented as a tensor in $\{ -1, +1 \}^f$, are thus given as input to a full-precision linear layer that outputs the logits for real/fake classification.

\section{Experiments}
\label{sec:experiments}

\subsection{Datasets}
\label{sec:datasets}
We leverage three distinct datasets, each offering unique challenges and characteristics, to effectively evaluate our deepfake detection method. 
The datasets are the COCOFake \cite{coco_fake}, the Diverse Fake Face Dataset (DFFD) \cite{dffd}, and the CIFAKE \cite{cifake}.

The COCOFake dataset builds upon the COCO dataset \cite{coco} and augments it with images using the Stable Diffusion \cite{stable_diffusion} text-to-image model. 
In particular, for each real image in COCO, which is accompanied by $5$ captions, $5$ corresponding synthetic images are created. 
This approach maintains the integrity of the original COCO dataset's division into training, validation, and test sets. 
The dataset comprises more than $650$K training images, $30$K validation images, and $30$K test images.

The DFFD is an extensive collection of images aimed at enhancing the detection and localization of facial manipulations. It covers four primary types of facial manipulations: identity swaps, expression swaps, attribute manipulations, and entirely synthesized faces. Real face samples are sourced from the FFHQ \cite{stylegan}, CelebA \cite{celeba} datasets (which offer a wide range of demographic and quality diversity), and additional real images from the FaceForensics++ \cite{faceforensics} dataset. For fake samples, PGGAN \cite{pggan} and StyleGAN \cite{stylegan} are employed to create manipulated images in line with the $4$ manipulation categories, summing up to $58$K real and $240$K fake images. Half of the samples are used in the training set, while $5\%$ for validation and $45\%$ for test sets, respectively, ensuring that manipulations derived from the same source image remain within the same set.

The CIFAKE dataset is structured to parallel the CIFAR-10 \cite{cifar} dataset, featuring a balanced composition of real and synthetic $32 \times 32$ pixels images across ten classes. Specifically, it includes $60$K real images directly taken from CIFAR-10 and an equal number of synthetic images generated using the Stable Diffusion \cite{stable_diffusion} model, summing up to $120$K images. The dataset is split into training and test sets, with $50$K images designated for the first and $10$K reserved for the latter.

\subsection{Metrics}
\label{sec:metrics}
Several metrics have been employed when comparing our method with the state of the art. These include metrics related to classification performance (accuracy, Area Under the Curve (AUC)), and computational requirements (floating point operations per second (FLOPs).
These metrics represent the standard in the deepfake detection \cite{coco_fake, dffd, cifake} and Binary Neural Networks \cite{lin2020rotated, tu2022adabin, guo2022join} fields.

Accuracy is defined as the ratio of correctly predicted observations, namely True Positives (TP) and True Negatives (TN), to the total observations, which includes False Positives (FP) and False Negatives (FN).
It is a measure of the model's overall correctness across all classes and is particularly useful for balanced datasets.

The AUC measures the ability of a model to discriminate between positive and negative classes. 
The AUC is the area under the Receiver Operating Characteristic (ROC) curve, which plots the TP rate against the FP rate at various thresholds.


The count of FLOPs is a measure of the computational complexity of a model, indicating the total number of floating point precision required for a single forward pass. 
This metric is crucial for understanding neural networks' computational demand and efficiency, especially when deploying models in resource-constrained environments.

\subsection{Experimental setup and training details}
\label{sec:experimental_setup}
Following the standard methodology used in related literature regarding the preprocessing of images, an initial resizing step was undertaken to standardize the longest dimension of each image to $252$ pixels. Subsequently, a central crop measuring $224\times224$ pixels was extracted from images within the validation and test sets. In contrast, crops from the training set were randomly obtained to introduce some variability. 
Data augmentation techniques were employed to augment the training dataset's diversity further. These techniques included random flips (both horizontal and vertical), rotations (either $90\degree$ or $270\degree$), and color jittering within a range of $[0\%, 20\%]$.
All images are then normalized to have mean $[0.485, 0.456, 0.406]$ and standard deviation $[0.229, 0.224, 0.225]$.
This particular choice of image size and color normalization has been done to maintain continuity with that of the data used during the pre-training of the backbone.

The model optimization was achieved using the AdamW \cite{adamw} optimizer in combination with the binary cross-entropy loss. 
It was configured with an initial learning rate of $10^{-4}$, with the first and second-moment estimates ($\beta_1$ and $\beta_2$) set to $0.9$ and $0.999$, respectively, accompanied by a weight decay parameter of $10^{-2}$. Additionally, a learning rate scheduler was implemented to methodically reduce the learning rate to $10^{-5}$ by the conclusion of the fifth epoch, optimizing the training process over time.
The batch size was set to $128$.
We experimented with two configurations of the models to evaluate their performance under different computational constraints: a version where the backbone is kept frozen to assess the model's behavior in conserving computational resources during training a trainable backbone to maximize the accuracy.
Based on the involved dataset, a maximum epoch limit was established throughout the experimental phase. For COCOFake and DFFD datasets (which details are provided in Section \ref{sec:datasets}) the epoch limit was set at $5$, a decision driven by the observation that the model convergence was typically achieved well before this threshold. Conversely, for the CIFAKE dataset, which is notably smaller in size, the epoch limit was extended to $20$ to accommodate the dataset's unique characteristics and ensure adequate model training.

The code was implemented in Python, leveraging the PyTorch framework for Deep Learning. Computational tasks were performed on an NVIDIA RTX 2080Ti GPU with 12GB of VRAM.

\subsection{Results}

\begin{table*}
	\centering
	\begin{tabular}{lll|cc|cc} 
\toprule
Method & Model & Pre-training dataset & Accuracy & AUC & Parameters (M) & FLOPs (G) \\ 
\midrule
\multirow{6}{*}{\cite{coco_fake}} & ResNet50 & ImageNet & 90.31 & - & 25.6 & 4.8 \\
 & ViT-B/32 & ImageNet & 87.64 & - & 88.3 & 8.56 \\
 & CLIP-ResNet50 & OpenAI WIT & 99.07 & - & 25.6 & 4.8 \\
 & CLIP-ViT-B/32 & OpenAI WIT & 99.11 & - & 88.3 & 8.56 \\
 & OpenCLIP-ViT-B/32 & LAION-400M & 97.88 & - & 88.3 & 8.56 \\
 & OpenCLIP-ViT-B/32 & LAION-2B & \textbf{99.68} & - & 88.3 & 8.56 \\ 
\hline
\multirow{6}{*}{Ours} & BNext-T with frozen backbone & ImageNet & 83.65 & 81.98 & 29.8 & \textbf{0.89} \\
 & BNext-S with frozen backbone & ImageNet & 93.15 & 95.19 & 67.1 & \underline{1.91} \\
 & BNext-M with frozen backbone & ImageNet & 84.59 & 82.11 & 133 & 3.39 \\
 & BNext-T & ImageNet & 99.25 & 99.86 & 29.8 & \textbf{0.89} \\
 & BNext-S & ImageNet & \underline{99.28} & \underline{99.89} & 67.1 & \underline{1.91} \\
 & BNext-M & ImageNet & 99.18 & \textbf{99.91} & 133 & 3.39 \\
\bottomrule
\end{tabular}
	\caption{Results on the COCOFake validation set. The models from \cite{coco_fake} were trained on different datasets, including ImageNet \cite{krizhevsky2012imagenet}, OpenAI WIT \cite{wit}, LAION-400M \cite{laion_400m} and LAION-2B \cite{laion_2b}. \textbf{Bold} and \underline{underlined} values respectively indicate the first and second-best results within their column.}
	\label{tab:results_coco_fake}
\end{table*}
\begin{table*}
	\centering
\begin{tabular}{ll|cc|cc} 
\toprule
Method & Model & Accuracy & AUC & Parameters (M) & FLOPs (G) \\ 
\midrule
\multirow{2}{*}{\cite{dffd}} & Xception & - & 99.64 & 40.0 & 18.0 \\ 
& VGG16 & - & 99.67 & 138.4 & 15.5 \\ 
\hline
\multirow{6}{*}{Ours} & BNext-T with frozen backbone & 89.56 & 87.65 & 29.8 & \textbf{0.89} \\
 & BNext-S with frozen backbone & 89.69 & 88.58 & 67.1 & \underline{1.91} \\
 & BNext-M with frozen backbone & 89.61 & 86.64 & 133 & 3.39 \\
 & BNext-T & \underline{98.95} & \textbf{99.94} & 29.8 & \textbf{0.89} \\
 & BNext-S & \textbf{99.01} & \textbf{99.94} & 67.1 & \underline{1.91} \\
 & BNext-M & 98.75 & \underline{99.92} & 133 & 3.39 \\
\bottomrule
\end{tabular}
	\caption{Results on the DFFD test set. \textbf{Bold} and \underline{underlined} values respectively indicate the first and second-best results within their column.}
	\label{tab:results_dffd}
\end{table*}
\begin{table*}
	\centering
	\begin{tabular}{@{}ll|cc|cc@{}}
		\toprule
		Method & Model & Accuracy & AUC & Parameters (M) & FLOPs (G) \\
		\midrule
		\multirow{3}{*}{\cite{wang2024harnessing}} & ResNet-50 & 95.00 & 99.00 & 25.6 & 4.8 \\
		   & VGG & 96.00 & 99.00 & 133 & 7.63 \\
		   & DenseNet & \textbf{98.00} & 99.00 & 7.9 & 5.6 \\
		\hline
		\multirow{6}{*}{Ours} & BNext-T with frozen backbone & 83.89 & 91.70 & 29.8 & \textbf{0.89} \\	
            & BNext-S with frozen backbone & 80.71 & 89.25 & 67.1 & \underline{1.91} \\	
            & BNext-M with frozen backbone & 82.77 & 90.73 & 133 & 3.39 \\	
            & BNext-T & 97.29 & \textbf{99.65} & 29.8 & \textbf{0.89} \\	
            & BNext-S & 96.96 & 99.55 & 67.1 & \underline{1.91} \\	
            & BNext-M & \underline{97.35} & \underline{99.62} & 133 & 3.39 \\	
		\bottomrule
	\end{tabular}
	\caption{Results on the CIFAKE test set. All the models were pre-trained on the ImageNet \cite{krizhevsky2012imagenet} dataset. \textbf{Bold} and \underline{underlined} values respectively indicate the first and second-best results within their column.}
	\label{tab:results_cifake}
\end{table*}

We report the results of our benchmark on the COCOFake dataset in Table \ref{tab:results_coco_fake}.
Our results were compared with the method proposed in \cite{coco_fake}.
To maintain a fair comparison, we set our model with ResNet-50 \cite{he2016deep} and ViT-B/32 \cite{vit}, both pre-trained on ImageNet as the models we used.
In the case of our models with a frozen backbone, we surpass the result of ResNet-50 by $2.84$ accuracy points with the table's second-best model, BNext-S.
When we also train the backbone, the margin of outperformance over ResNet-50 expanded to $8.97$ accuracy points.
This proves that our model can perform better than full-precision models initialized on the same dataset while having substantially lower FLOPs.
Furthermore, our approach remains competitive when comparing our method with models pre-trained on substantially larger datasets.
With its $99.28\%$ accuracy, our best model trails the best-performing model, OpenCLIP-ViT-B/32 trained on LAION-2B, by just $0.4$ points.
Notably, the latter model was trained on a dataset comprising $2$ billion images, in contrast to the $1.2$ million images of the ImageNet dataset in which our models were pre-trained; this fact highlights the efficacy of our method despite its reduced precision.

The outcomes on the DFFD dataset, detailed in Table \ref{tab:results_dffd}, are compared against the performances of Xception \cite{xception} and VGG16 \cite{vgg} as outlined in \cite{dffd}. 
As noticed, our models with a trainable backbone consistently perform better than the two compared models, achieving an improvement of $0.27$ AUC points while having up to $17.1$ fewer GFLOPs.

Regarding the results on the CIFAKE dataset, shown in Table \ref{tab:results_cifake}, we compare our method with the models proposed in \cite{cifake}.
To keep a fair comparison, we consider the results of BNext with a trainable backbone given the same training conditions.
However, \cite{cifake} does not specify the exact version of VGG and DenseNet \cite{densenet}, making it impossible to directly compare the number of parameters and FLOPs. 
To aid the reader, we have included metrics for the smallest versions of these models. 
Despite this, BNext demonstrates competitive performance while requiring fewer GFLOPs.

\subsection{Ablation study}
We conducted an ablation study to ascertain the optimal amalgamation of features incorporated into the input. The results are delineated in Table \ref{tab:ablation}.
Specifically, this study juxtaposed a baseline model against various configurations incorporating supplementary channels, using accuracy as a metric since the differences in the number of FLOPs are negligible.
The baseline model processes an RGB image as input, directly channeling it into a pre-trained BNext-T backbone without integrating an Adapter. Conversely, the alternative models evaluated entail baseline variations, each retrained with the inclusion of one or more of the additional channels, each paired with a congruent Adapter.
These new channels are the FFT magnitude and LBP described in Section \ref{sec:method}, plus the one obtained by applying a Sobel filter to the image.
The latter is employed to accentuate the edge information of the image and is obtained by applying a pair of $3 \times 3$ convolution kernels, one estimating the gradient horizontally and the other vertically, to approximate the gradient magnitude of the image at each point. 
By emphasizing edges and contours, we thought the Sobel filter would aid in highlighting discrepancies in the boundary regions often overlooked by deepfakes, focusing on the premise that genuine images possess naturally smooth transitions, which manipulated images struggle to replicate accurately.

The ablation study highlights that FFT magnitude and LBP channels, when combined, markedly improve model performance in detecting manipulated images, surpassing the baseline and other variations.
The increase amounts to $1.25\%$ over the second best-performing configuration, which is the baseline.
This synergy stems from their complementary analytical approaches: FFT magnitude exposes anomalies in the frequency domain indicative of digital manipulation. At the same time, LBP captures nuanced local texture patterns disrupted by such manipulations. 
Alone, each channel is seen to have a partial view: FFT magnitude might miss subtle textural alterations, and LBP could overlook frequency-based distortions. 
Together, they cover spectral and spatial discrepancies, enhancing detection capabilities. 
The Sobel filter's marginal impact suggests that edge information alone is insufficient for deepfake detection, underscoring the importance of integrating features that address global and local image characteristics for optimal performance.

\begin{table}
	\centering
	\begin{tabular}{@{}l|l|cc|cc@{}}
		\toprule
		Ablation                               & Variations                                & Accuracy (\%) \\
		\midrule
		Baseline                               & -                                         & \underline{90.35} \\
		\hline
		\multirow{3}{*}{\shortstack[l]{Features added \\ to the \\ learned ones}} & Magnitude & 82.36 \\
		                                       & FFT                                & 88.18 \\
		                                       & LBP & 88.42 \\
                                         & Magnitude and FFT & 81.20 \\
                                         & Magnitude and LBP & 81.67 \\
                                         & FFT and LBP & \textbf{91.60} \\
                                         & Magnitude, FFT and LBP & 81.56 \\
				
		\bottomrule
	\end{tabular}
	\caption{Ablation study on the COCOFake Dataset.}
	\label{tab:ablation}
\end{table}  
\section{Conclusion}
\label{sec:conclusion}
In this study, we investigate the performance of more computationally efficient neural networks, particularly BNNs, in the context of deepfake detection tasks. Our findings reveal that the proposed BNN-based method, which requires up to $5$ times fewer FLOPs compared to a ResNet-50 model and nearly $10$ times fewer FLOPs than a ViT-B/32 model, is capable of matching the performance of their full-precision counterparts with minimal loss in classification accuracy. These results suggest a promising direction for enhancing the efficiency of deepfake detection methodologies.

One notable limitation of our study is the emphasis on theoretical FLOP reductions, as the real-world application of BNNs necessitates a specialized framework or accelerator to fully realize the benefits of reduced precision. Furthermore, our evaluation was confined to a network pre-trained on the ImageNet dataset, whereas other investigations have leveraged larger datasets for pre-training, thereby achieving enhanced transfer-learning capabilities.

For future research, there is potential for practical implementation of our proposed method on specialized hardware or within specific computational frameworks to actualize the theoretical efficiency gains. Additionally, exploring alternative pre-training datasets could further augment the transfer-learning efficacy of the network, potentially leading to more robust and efficient deepfake detection systems.  

\section{Acknoledgements}

The research leading to these results has received funding from Project “Ecosistema dell’innovazione - Rome Technopole” financed by EU in NextGenerationEU plan through MUR Decree n. 1051 23.06.2022 - CUP H33C22000420001.

{
    \small
    \bibliographystyle{ieeenat_fullname}
    \bibliography{main}
}

\end{document}